\documentclass{pprai}

\usepackage[utf8]{inputenc}
\usepackage[T1]{fontenc}
\usepackage{graphicx}
\usepackage{txfonts}
\usepackage{url}
\usepackage{algpseudocode}
\usepackage{subfigure} % for adding two images in the same row next to each other
\usepackage{booktabs} % for table with results
\usepackage{siunitx} % Required for alignment
\usepackage{makecell} % for making cells in table plots in summary
\usepackage{hyperref}
\usepackage{wrapfig}
\usepackage[export]{adjustbox}

\title{Deep Generative Models for Proton Zero Degree Calorimeter Simulations in ALICE, CERN}
\headtitle{Deep Generative Models for Proton ZDC Simulations in ALICE, CERN}

\author{Patryk Będkowski$^{1}$, Jan Dubiński$^{1, 2}$, Kamil Deja$^{1, 2}$, Przemysław Rokita$^{1}$}
\headauthor{Patryk Będkowski et al.}
\affiliation{%
  $^1$Warsaw University of Technology\\
  $^2$IDEAS NCBR}


\begin{document}
\maketitle

\begin{abstract}

Simulating detector responses is a crucial part of understanding the inner-workings of particle collisions in the Large Hadron Collider at CERN. The current reliance on statistical Monte-Carlo simulations strains CERN's computational grid, underscoring the urgency for more efficient alternatives. Addressing these challenges, recent proposals advocate for generative machine learning methods. In this study, we present an innovative deep learning simulation approach tailored for the proton Zero Degree Calorimeter in the ALICE experiment. Leveraging a Generative Adversarial Network model with Selective Diversity Increase loss, we directly simulate calorimeter responses. To enhance its capabilities in modeling a broad range of calorimeter response intensities, we expand the SDI-GAN architecture with additional regularization. Moreover, to improve the spatial fidelity of the generated data, we introduce an auxiliary regressor network. Our method offers a significant speedup when comparing to the traditional Monte-Carlo-based approaches.
%reduction in the computational complexity and 

\end{abstract}

\section{Introduction}

ALICE (A Large Ion Collider Experiment) is one of the four major detectors located at the Large Hadron Collider at CERN. 
One of its main goals is to replicate and study the intense conditions that existed in the early universe shortly after the Big Bang. Apart from gathering real data from the experiment, in order to understand the properties and events of particle collisions scientists have to perform complex simulations that compared with experimental data can validate hypotheses. Such simulations are computationally extremely expensive. Existing approaches utilize statistical Monte-Carlo methods to model the physical interactions between particles. While these methods yield high-fidelity outcomes, they are also associated with high computational demands. In 2023, over 540'000 CPU devices \cite{lhccomputationreport} were engaged in the computations of ALICE experiments, marking a demand for developing more efficient simulation methods.

One of the most computationally expensive part of the process is the simulation of Calorimeter. Therefore, in this study, we focus on machine learning models to simulate data from the Proton Zero Degree Calorimeter (ZDC) of the ALICE experiment at CERN. We implement a deep convolutional Generative Adversarial Network (GAN)~\cite{gan} as a baseline model and adapt the SDI-GAN model \cite{sdigan2023} which incorporates a regularization technique aimed at increasing the diversity of generated samples. Then, we extend the model by a simple, yet powerful regularization method focused on minimizing the difference in intensities between real and generated calorimeter responses, which improves the quality of the simulations. Finally, we introduce an auxiliary regressor which increases the model capabilities to learn accurate spatial features of the simulation data. We can summarise the contributions of this work as follows:
\begin{itemize}
    \vspace{-0.2cm}
    \item We develop the first generative simulation method for the proton Zero Degree Calorimeter in the ALICE experiment at CERN.
    \vspace{-0.2cm}
    \item We evaluate the baseline GAN and SDI-GAN in our simulation task.
    \vspace{-0.2cm}
    \item We extend the SDI-GAN model by an intensity regularization loss and spatial auxiliary regressor, achieving increased simulation quality.
\end{itemize}

\section{Related work}
Throughout recent years, the use of Generative AI for various CERN simulations has shown the versatility of these methods~\cite{deja2018generative,deja2020endtoend,sdigan2023,Dubinski2023MachineLM}. 
Authors of \cite{Dubinski2023MachineLM} employ generative machine learning algorithms for the task of simulating a Neutron ZDC calorimeter device. They propose a solution that utilizes generative models, specifically focusing on the performance of variational autoencoders and generative adversarial networks. By expanding the GAN architecture with an additional regularization, the authors significantly increase the simulation speed by two orders of magnitude while maintaining the high fidelity of the simulation.

To increase the diversity of the simulations present in the dataset, in \cite{sdigan2023} authors present a model dubbed SDI-GAN which offers significant improvements to the simulations by enforcing sample diversity among subsets of conditional data without affecting samples that exhibit consistent responses.

Finally, in order to improve how the generator learns the geometric properties of the data, in \cite{Chekalina:2018hxi, gan-aux-cern} authors employ a regressor to accurately align the shower's center in calorimeter responses. In this work, we introduce a similar approach in the ZDC Calorimeter in ALICE.

\section{Method}

We select a Deep Convolutional Generative Adversarial Network \cite{gan} as our experimental model, consisting of a generator $G(z)$ and a discriminator $D(x)$. The generator creates image $x$ from random noise $z$, and the discriminator differentiates between real and generated images. Both networks are conditioned on particle data and trained adversarially. Post-training, the generator is used to synthesize new calorimeter responses to particle collisions. During training, given conditional input $c$ and $k$-dimensional latent code $z \sim \mathcal{N}_k(0, 1)$, generator $G(z, c)$ produces an output image $\hat{x} = G(z, c)$.
\setlength{\abovedisplayskip}{6pt} % Adjust the space above the equation
\begin{equation}
L_{adv}(G, D) = \mathbb{E}_{x \sim \mathcal{X}, c \sim \mathcal{C}}[\log D(x, c)] + \mathbb{E}_{c \sim \mathcal{C}, z \sim \mathcal{N}(0,1)}[\log (1 - D(G(c, z), c))]
\end{equation}

\begin{figure}[!h]
	\label{fig:gen-architecture}
	\centering \includegraphics[width=0.75\linewidth]{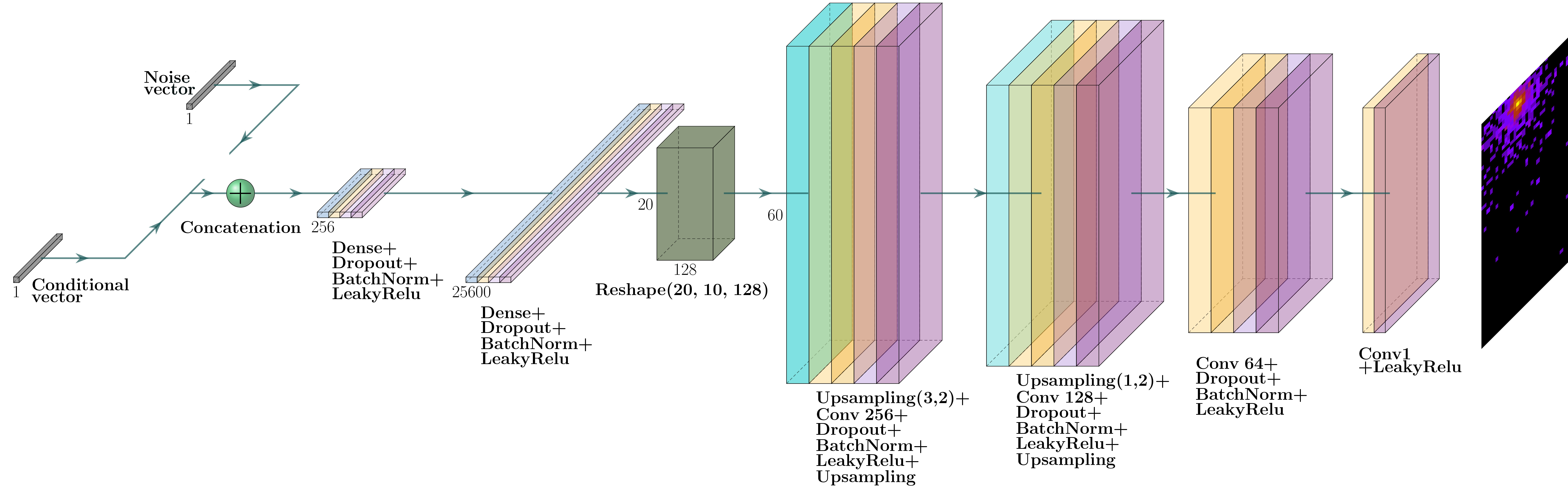}
 	\caption{\small Architecture of the generator in used convolutional GAN across all tests.}
\end{figure}

\subsection{Auxiliary regressor}

One of the main properties of calorimeter response is the localisation of the center of the shower -- area where the pixel values are the highest. Therefore, to improve the GAN's learning of geometric data properties, we introduce an auxiliary regressor that trains alongside the main network to identify the 2D coordinates of the center of the collision. In preprocessing, we calculate these coordinates for all training samples, using them as targets for training. The auxiliary loss, measured by the mean squared error between the predicted $(\hat{k_{i}}, \hat{l_{i}})$ and actual coordinates $(k_{i}, l_{i})$ of the highest-value pixel of the image $x_i$, is added to the generator's loss to refine its geometric accuracy. Strength is controlled by $\lambda_{aux}$ parameter.

\begin{equation}
L_{aux} = \frac{1}{N} \sum_{i=1}^{N} \left[ (\hat{k_{i}} - k_{i})^2 + (\hat{l_{i}} - l_{i})^2 \right]
\end{equation}

\subsection{Diversity regularization}

Authors of SDI-GAN propose a regularization which seeks to minimize the ratio between the $L1$ distance of two images $d_I$ generated from two distinct latent codes $z_1$, $z_2$ under the same conditioning vector \textit{c} and the $L1$ distance between the latent codes themselves $d_z$. The measure of diversity is based on the variance in the original dataset. Thus, as a preprocessing for each unique conditioning value $\textit{c}$, they calculate the variance of pixel values among samples. Later obtained values of all sample diversity are normalized to \([0, 1]\) by dividing by the length of dataset. This diversity measure is then multiplied with the regularization term $\lambda_{div}$, adjusting its influence on training objectives.

\begin{equation}
    L_{div} = \sum_{i, j} \sqrt{\frac{\sum_{t}(x_{ij}^{t} - \mu_{ij})^2}{|X|}} \times \left(\frac{d_I (G(c, z_1), G(c, z_2))}{d_z (z_1, z_2)} \right)^{-1}
\end{equation}
where $i$ and $j$ are the pixel coordinates, $t$ is the index of sample $x$ $\in$ \text{\Large $\chi$} and $\mu_{ij}$ is a mean value for a pixel $ij$.

\subsection{Intensity regularization}
SDI-GAN performs well on filtered data but struggles with varying intensity levels of Cherenkov light particles across distributions. To address this, we introduce a regularization using intensity measure $f_{in}$ from the original dataset. In preprocessing, for each conditional vector $\textit{c}$, we calculate this intensity measure as the pixel sum of the corresponding image $x$ from $\text{\Large $\chi$}$
$f_{in}(x) = \sum_{i, j}x_{ij}$, where \textit{i}, \textit{j} are coordinates of an image $x$. The intensity between the generated image $\hat{x}$ and the original sample $x$ corresponding to $c$ is calculated using the mean absolute error (MAE). This loss is then multiplied by the constant $\lambda_{in}$ to adjust its strength.

\begin{equation}
    L_{in} = |f_{in}(x_c) - f_{in}(\hat{x}_c)|
\end{equation}

\subsection{Final training objective}
The final training loss for our approach is a combination of the basic GAN training loss with the proposed modifications, and can be summarised as follows:

\begin{equation}
    L(G, D) = L_{adv}(G, D) + \lambda_{div}L_{div}(\textit{G}) + \lambda_{in}L_{in}(\textit{G}) + \lambda_{aux}L_{aux}
\end{equation}

\section{Experiments}

\begin{wrapfigure}{r}{0.5\linewidth}
\vskip  -16pt
    \begin{adjustbox}{center}
    \includegraphics[height=1.3\linewidth]{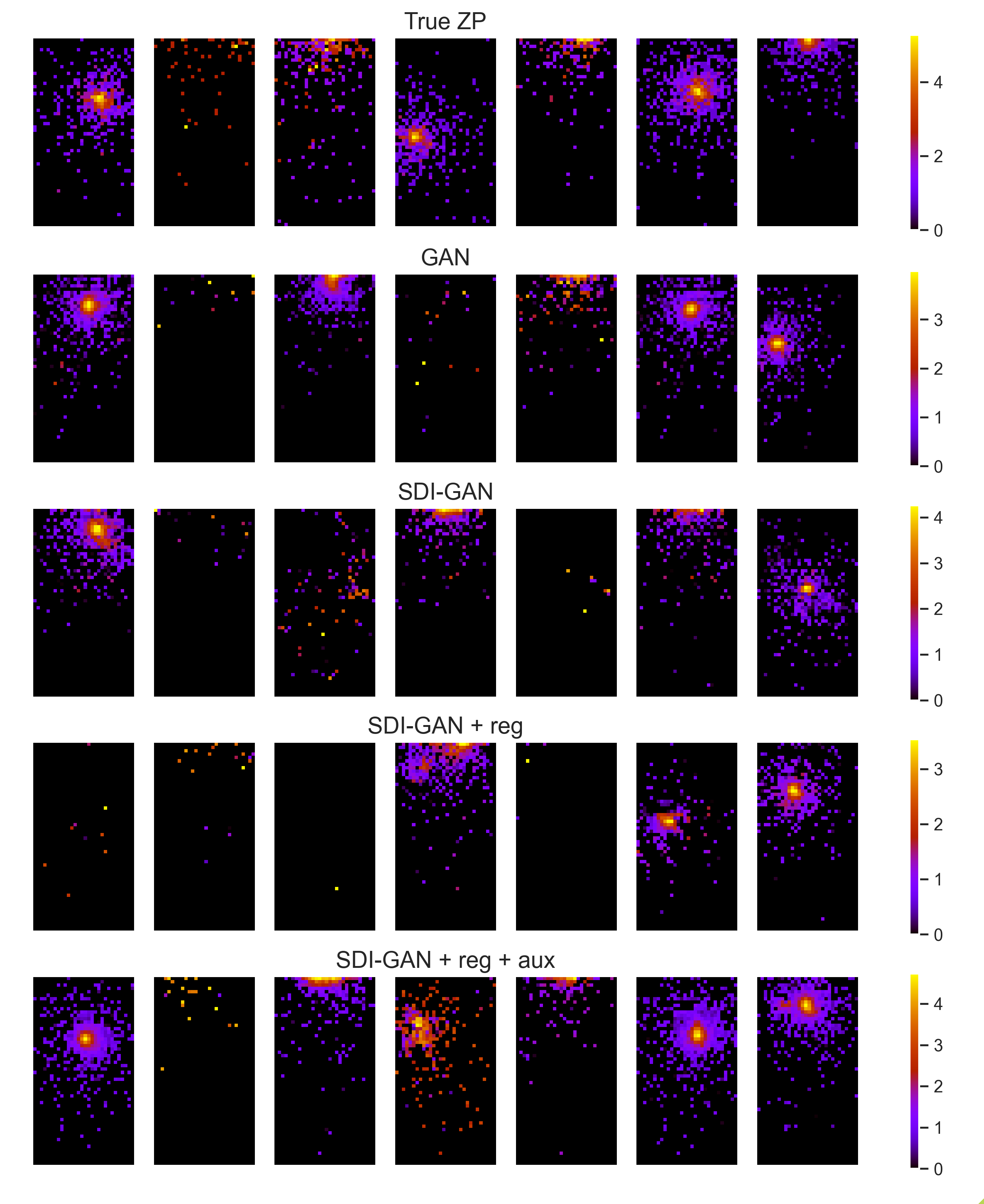}
    \end{adjustbox}
    \vskip -12pt
    \caption{\small \label{fig:example-responses} Example of simulated responses from different methods.}
\vskip  -30pt
    % \end{figure}
\end{wrapfigure}

The Zero Degree Calorimeter (ZDC) includes a Proton (ZP) device for recording energy from non-interacting nuclei in collisions. It uses silica fibers to detect Cherenkov light, converting it into 2D images with a resolution of 56 by 30 pixels. The visualisation of the original detector responses is presented in Fig.~\ref{fig:example-responses} (top row). Each simulation example is therefore an image coming from a ZP device which is referred to as a response of the experiment associated with a vector of variables, referred to as conditional data. Conditional data compromises of 9 variables: energy, mass, charge, three spatial position coordinates, and three momentum coordinates. 

\begin{table}[b]
\small
% \vspace{-20pt}
\centering
\caption{\small \label{tab:results}Comparison of mean WS metric across five runs.}
\label{ws-distance-comparison-1}
\begin{tabular}{p{6cm}p{2.3cm}p{2.3cm}}
\toprule
Model & {WS Distance $\downarrow$} & {Std Dev} \\
\midrule
GAN & 2.4752 & 1.6843 \\
SDI-GAN & 2.3571 & 1.6000 \\
SDI-GAN + intensity reg. & 2.2916 & 1.8210  \\
SDI-GAN + intensity reg. + aux. reg. & \textbf{2.0777} & 1.6381\\
\bottomrule
\end{tabular}
% \vspace{-1pt}
\label{ws-distance-comparison}
\end{table}

\newpage
The dataset employed in this experiment constitutes almost 350 thousand samples, with the validation method incorporating an 80:20 train-test split ratio. Our evaluation methodology is founded on the analysis of five distinct channels outlined in the calorimeter's specifications \cite{alice-tech-report}. We employ the standard 1st Wasserstein distance metric \cite{ws-dist} to assess the fidelity of the simulations across all channels. As shown in Tab.~\ref{tab:results}, our modifications positively influence the quality of generations as measured by the Wasserstein distance.

\begin{figure}[!h]
\vspace{-0.1cm}
	\label{fig:distributions}
	\centering \includegraphics[width=1\linewidth]{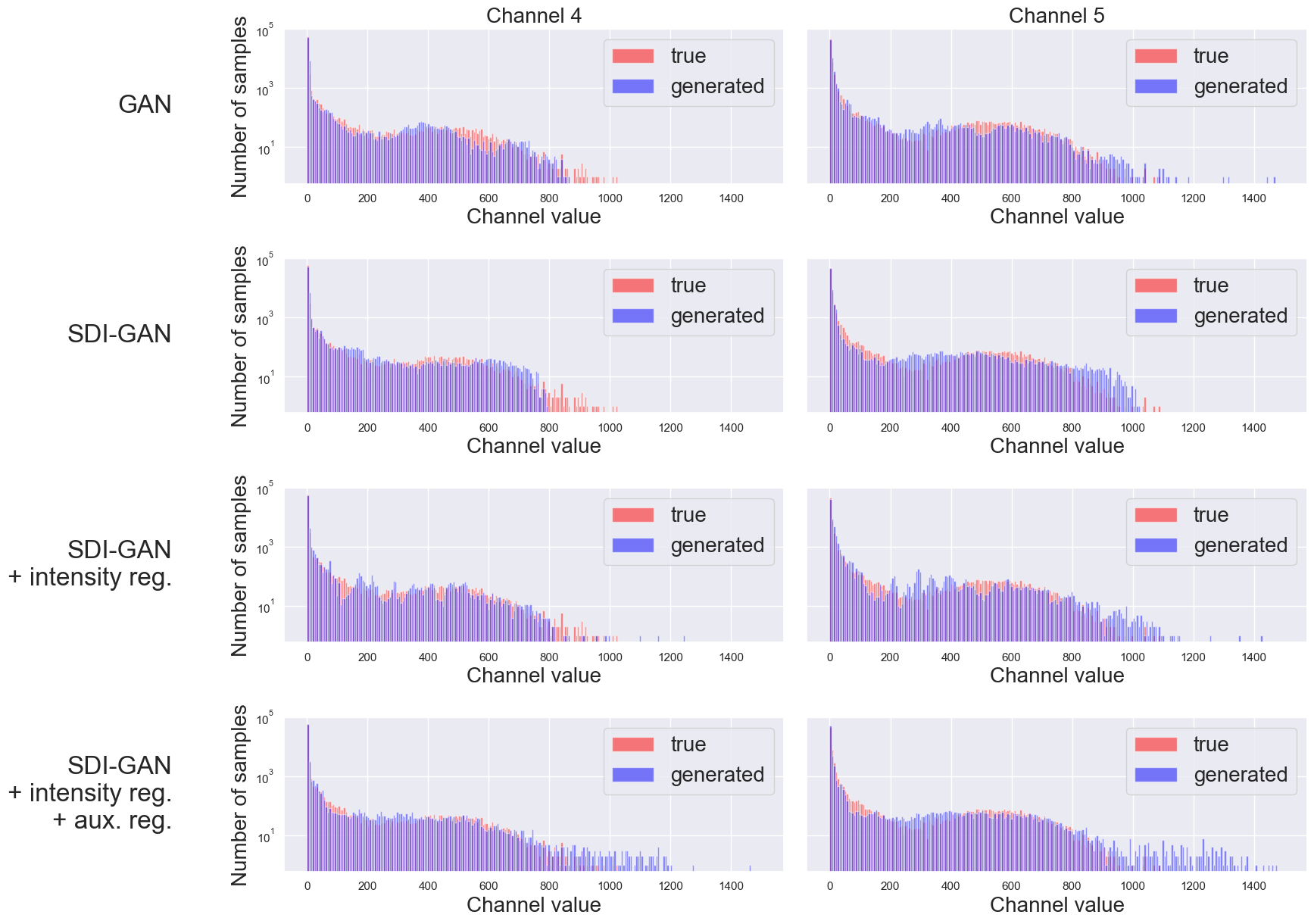}
 	\caption{\small Histograms of true and generated distributions of channel values. The GAN and SDI-GAN model have visible problems with underproducing high-energy responses. %The SDI-GAN does not completely allow the model to fit the real distribution. 
  The implementation of additional regularization, and auxiliary regressor positively influence better alignment to true distribution, but tends to oversample the high-energy responses.
 }
 \vspace{-1pt}
\end{figure}

To visualise the differences between different methods, we plot responses to several different parameters in Fig.~\ref{fig:example-responses}. The basic GAN produces images that are not visually aligned with the actual data. Samples generated by SDI-GAN struggle to fit the positions of real samples. Auxiliary regressor better aligns the positions of centers of particle showers. In Fig.~\ref{fig:distributions} we present the visual results for channel 4 and 5, as they contain most of information.

\clearpage
\section{Conclusions}
In this study, we applied and extended generative machine learning models to simulate the intricate dynamics of the Proton Zero Degree Calorimeter in the ALICE experiment at CERN. By establishing GAN as a baseline model we tested its performance in comparison to the SDI-GAN. Furthermore, we tailored SDI-GAN to the unique demands of High Energy Physics experiments, by adding additional regularization on the sum of pixel intensities which proved to increase the fidelity of the simulations. Further, the incorporation of an auxiliary regressor proved essential to securing the lowest WS results.

\section*{Acknowledgments}

This research was funded by National Science Centre, Poland grants: 2020/39/ O/ST6/01478 and 2022/45/B/ST6/02817. This research was supported in part by PLGrid Infrastructure grants: PLG/2023/016393, PLG/2023/016361, PLG/2023/ 016278.

\section*{Appendix: training details}

Fine-tuning was conducted via grid search across logarithmically spaced values: \( \lambda_{div} \in \{10^{-2}, 10^{-1}, 10^{0}\} \), \( \lambda_{in} \in \{10^{-7}, 10^{-8}, 10^{-9}, 10^{-10}, 10^{-11}\} \), and \( \lambda_{aux} \in \{10^{-4}, 10^{-3}, 10^{-2}\} \). Model configurations were assessed for the lowest WS over five runs, with the best performance yielded by \( \lambda_{div} = 10^{-1} \), \( \lambda_{in} = 10^{-10} \), and \( \lambda_{aux} = 10^{-3} \). The rest of the parameters were constant for all tests.

\clearpage
\bibliography{pprai}
\bibliographystyle{pprai}

\end{document}